\documentclass[sigconf, nonacm]{acmart}
\AtBeginDocument{%
  \providecommand\BibTeX{{%
    Bib\TeX}}}

\usepackage{enumitem}
\usepackage{newunicodechar} 
\usepackage{algorithmic}
\usepackage{graphicx}
\usepackage{textcomp}
\usepackage{xcolor}
\usepackage[table]{xcolor}
\usepackage{multirow}
\usepackage{booktabs}
\usepackage{subcaption}    
\usepackage[ruled,vlined]{algorithm2e}
\def\BibTeX{{\rm B\kern-.05em{\sc i\kern-.025em b}\kern-.08em
    T\kern-.1667em\lower.7ex\hbox{E}\kern-.125emX}}
\begin{document}

\title{On-Demand Multi-Task Sparsity for Efficient Large-Model Deployment on Edge Devices
}
\author{Lianming Huang}
\email{lmhuang8-c@my.cityu.edu.hk}
\affiliation{%
  \institution{City University of Hongkong}
    \city{Hong Kong}
  \country{China}
}
\author{Haibo Hu}
\email{haibohu2-c@my.cityu.edu.hk}
\affiliation{%
  \institution{City University of Hongkong}
  \city{Hong Kong}
  \country{China}
}
\author{Qiao Li}
\email{qiaoli045@gmail.com}
\affiliation{%
  \institution{Mohamed bin Zayed University of Artificial Intelligence}
    \city{Abu Dhabi}
  \country{The United Arab Emirates}
}
\author{Nan Guan}
\email{nanguan@cityu.edu.hk}
\affiliation{%
  \institution{City University of Hongkong}
    \city{Hong Kong}
  \country{China}
}
\author{Chun Jason Xue}
\email{jason.xue@mbzuai.ac.ae}
\affiliation{%
  \institution{Mohamed bin Zayed University of Artificial Intelligence}
    \city{Abu Dhabi}
  \country{The United Arab Emirates}
}

\begin{abstract}
Sparsity is essential for deploying large models on resource constrained edge platforms. However, optimizing sparsity patterns for individual tasks in isolation ignores the significant I/O overhead incurred during frequent task switching. We introduce an on-demand multi-task sparsity framework specifically designed to minimize switching costs by maximizing parameter reuse. Unlike monolithic approaches, we decompose weights into reusable block-granular units and align sparse structures across tasks to maximize overlap. 
By dynamically loading only the small differential set of blocks required for the next task, our method effectively mitigates the cold-start latency inherent in traditional monolithic approaches.
Experiments on a real-world autonomous driving platform demonstrate that our framework achieves superior switching efficiency, accelerating task switching by over $6.6\times$ on average compared to existing sparsity methods.
\end{abstract}
\maketitle
\thanks{This is a preprint of a paper submitted to DAC2026.}
\keywords{Multi-Task Switch, Model Sparsity, Dynamic Weight Loading, Autonomous Driving, Vision–Language Models}

\section{Introduction}
Large-scale language and vision models have achieved remarkable performance in multimodal, language, and vision tasks~\cite{hoefler2021sparsity,liu2023deja,liu2024deepseek,cai2025survey,fedus2022switch}.
However, their massive parameter sizes and memory footprints make deployment difficult on resource-constrained edge platforms, such as in-vehicle GPUs and embedded accelerators~\cite{yu2024edge,kim2024monde,zhang2025memorianova}.
In autonomous driving and other safety-critical scenarios, a single system must support multiple perception and reasoning tasks, including traffic light recognition, obstacle detection, scene understanding, and language-guided reasoning, while meeting strict latency and reliability constraints.
Running separate full models for each task is impractical due to memory constraints in certain resource-constrained conditions.

Sparsity has become a central tool for improving the efficiency and conserving computational resources of large models.
Pruning and growth methods~\cite{hoefler2021sparsity,zheng2024learn,hou2025instruction,huang2023dynamic} compress dense networks by removing redundant parameters; and Mixture-of-Experts (MoE) architectures~\cite{liu2024deepseek,cai2025survey,liu2024survey,fedus2022switch} scale model capacity while routing each input through only a few experts; activation and contextual sparsity~\cite{liu2023deja,liu2024training,dhar2024activation,qi2025deltallm} selectively activate subsets of neurons or tokens at inference time.
These techniques substantially reduce computation and active parameters for a \emph{single} workload.
However, they typically assume a monolithic model: once a checkpoint is loaded, sparsity only controls which parameters are used for a specific task, ignoring how the weights are stored, shared, or swapped across multiple tasks.
When multiple tasks are multiplexed on a shared device, each with its own sparsity pattern, switching tasks still requires repeatedly fetching large and often disjoint parameter subsets from CPU memory to GPU, so host–device I/O during task switches becomes the dominant source of cold-start latency~\cite{dayan2025reducingcoldstart}.

Early-exit and skip-layer approaches provide another important dimension of efficiency.
Comprehensive surveys~\cite{rahmath2024early,bajpai2025survey} and recent methods~\cite{elhoushi2024layerskip,huang2025gm,sponner2024temporal,regol2023jointly,qi2025deltallm} demonstrate that many inputs can be confidently processed at shallow layers, enabling dynamic depth and self-speculative decoding in language models, as well as temporal reuse and joint exit policies in deep networks.
In the multimodal autonomous driving domain, navigation-guided early exiting~\cite{hu2025nav} and metric-guided block skipping~\cite{huang2025gm} have been shown to accelerate vision-language models while maintaining accuracy.
However, these methods largely focus on \emph{in-task} computation reduction. They choose when to stop or which blocks to skip for a given task, but do not coordinate sparse structures across heterogeneous tasks.
As a result, the dominant cost in multi-task edge deployment, i.e., moving parameters among storage, CPU memory, and GPU memory when tasks switch, is left unaddressed.

In this paper, we propose an on-demand multi-task sparsity framework that jointly considers model storage layout, sparsity structures, and task scheduling for large-model deployment on edge platforms.
First, we reorganize model weights into reusable units, enabling fine-grained and flexible management for large models. 
Second, rather than optimizing sparsity independently for each task, we align sparse patterns across tasks to maximize overlap, minimizing the switch costs.
Third, by analyzing the correlation among tasks, we pre-load a small set of parameters required by likely upcoming tasks.
Our design enables efficient multi-task switching with reduced memory footprint, lower I/O volume, and more stable real-time throughput, making sparsity techniques better suited for practical edge and in-vehicle deployments.
Empirically, on LLaVA-7B, our framework reduces the task-switching latency between traffic-light decision and vehicle detection, achieving a $9.8\times$ speedup on a real-world autonomous driving platform.

The main contributions of this work are summarized as follows:
\begin{itemize}[topsep=2pt,parsep=0pt,partopsep=0pt,itemsep=1pt]
  \item \textbf{On-demand, fine-grained weight loading.}
  We decompose model weights into reusable units and load only necessary blocks at inference time, reducing GPU memory usage and I/O compared to monolithic checkpoints.
  \item \textbf{Overlap-aware sparsity alignment.} We align task-specific sparse patterns to maximize shared active model blocks across tasks, reducing the amount of parameters that need to be updated on each task switch.
\item \textbf{Correlation-aware pre-loading.} We leverage task-transition statistics to proactively stage block layers associated with highly probable next tasks into CPU memory. This hides disk access latency during multi-task switches, improving responsiveness without increasing GPU load.
  \item \textbf{Edge and real-vehicle validation.} We deploy our framework with vision-language models for autonomous driving on an on-board edge device in a real vehicle, improving task-switching efficiency by more than $6.6\times$ on average while preserving task performance.
\end{itemize}

\section{Background and Motivation}
\label{sec:motivation}

\subsection{Background}

Sparsity is a key tool for making large models practical: pruning~\cite{hoefler2021sparsity,zheng2024learn,hou2025instruction} removes redundant weights, activation/contextual sparsity~\cite{liu2023deja,liu2024training,dhar2024activation,qi2025deltallm} activates only a small subset of neurons or tokens, MoE architectures~\cite{liu2024deepseek,cai2025survey,liu2024survey,fedus2022switch} route each token through a few experts, and early-exit/skip-layer techniques~\cite{rahmath2024early,elhoushi2024layerskip,hu2025nav,huang2025gm,bajpai2025survey,sponner2024temporal,regol2023jointly,qi2025deltallm} stop inference at shallow layers.  
However, these methods are largely tailored to a \emph{single} workload with a monolithic checkpoint: they select active parameters per input but ignore cross-task parameter storage and reuse, leaving task-switching costs unaddressed.

\begin{figure}[t]
  \centering
  \includegraphics[width=0.4\textwidth,trim=0cm 0cm 17cm 0cm, clip]{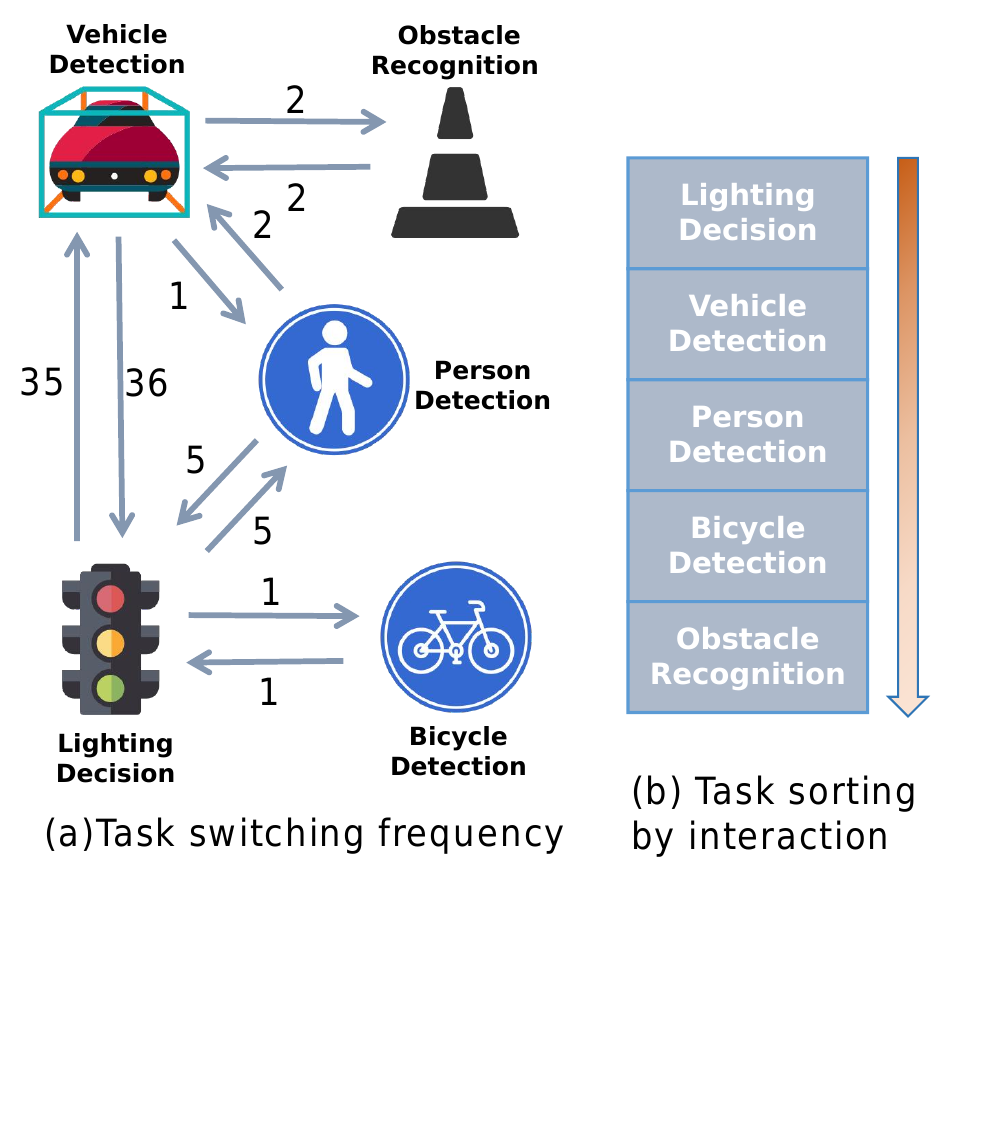}
  \vspace{-55pt}
  \caption{Task-switching frequency graph and sort, where arrow directions indicate switches between tasks and edge labels denote how often each transition occurs.}
  \vspace{-10pt} 
  \label{fig:switching}
\end{figure}

To understand the consequences in realistic workloads, we analyze task switching on the nuScenes trainval split over five perception modules: traffic-light decision, vehicle detection, obstacle recognition, person detection, and bicycle detection, shown in Fig.~\ref{fig:switching}.  
The counts show that bidirectional transitions between vehicle detection and lighting decision dominate, while switches involving obstacle, person, and bicycle modules are less frequent, yielding dense but highly skewed transitions among heterogeneous tasks with distinct sparsity and latency requirements.  
Under such workloads, keeping a large MoE fully resident quickly exhausts GPU memory, while deploying separate sparse models per task makes every switch expensive, as GPU-resident weights must be replaced by large parameter sets fetched from CPU.

\subsection{Motivation}
In MoE-style deployments, all experts are kept resident in GPU memory so that task switches are cheap, only the routing pattern changes, which is primarily suitable for cloud settings with abundant GPU resources.
Skip/EE-style deployments, in contrast, do achieve substantial parameter and resource reduction by learning task-specific skip patterns or exit policies. However, these sparse configurations are usually obtained via additional fine-tuning on each downstream task and do not transfer well across domains.
As a result, in multi-task scenarios, one typically maintains a separate sparse configuration per task and switches by replacing the entire set of active weights.
In contrast, we seek a deployment strategy where tasks share as many active parameters as possible, and where only a small differential set of weights needs to be switched when the task changes.

These observations necessitate a storage- and switching-aware view of sparsity in multi-task edge deployment.
Rather than optimizing sparsity separately for each task on top of a fixed storage layout, we aim to:
(i) reorganize model weights so that parameters can be fetched on demand at fine granularity;
(ii) align sparse structures across tasks to maximize parameter reuse during switches; and
(iii) exploit task-transition statistics to pre-load only a small set of high-priority parameters for likely upcoming tasks.

\begin{figure*}[!t]
\centering
\includegraphics[width=1\textwidth, height=0.4\textheight,trim=0cm 0cm 2cm 0cm]{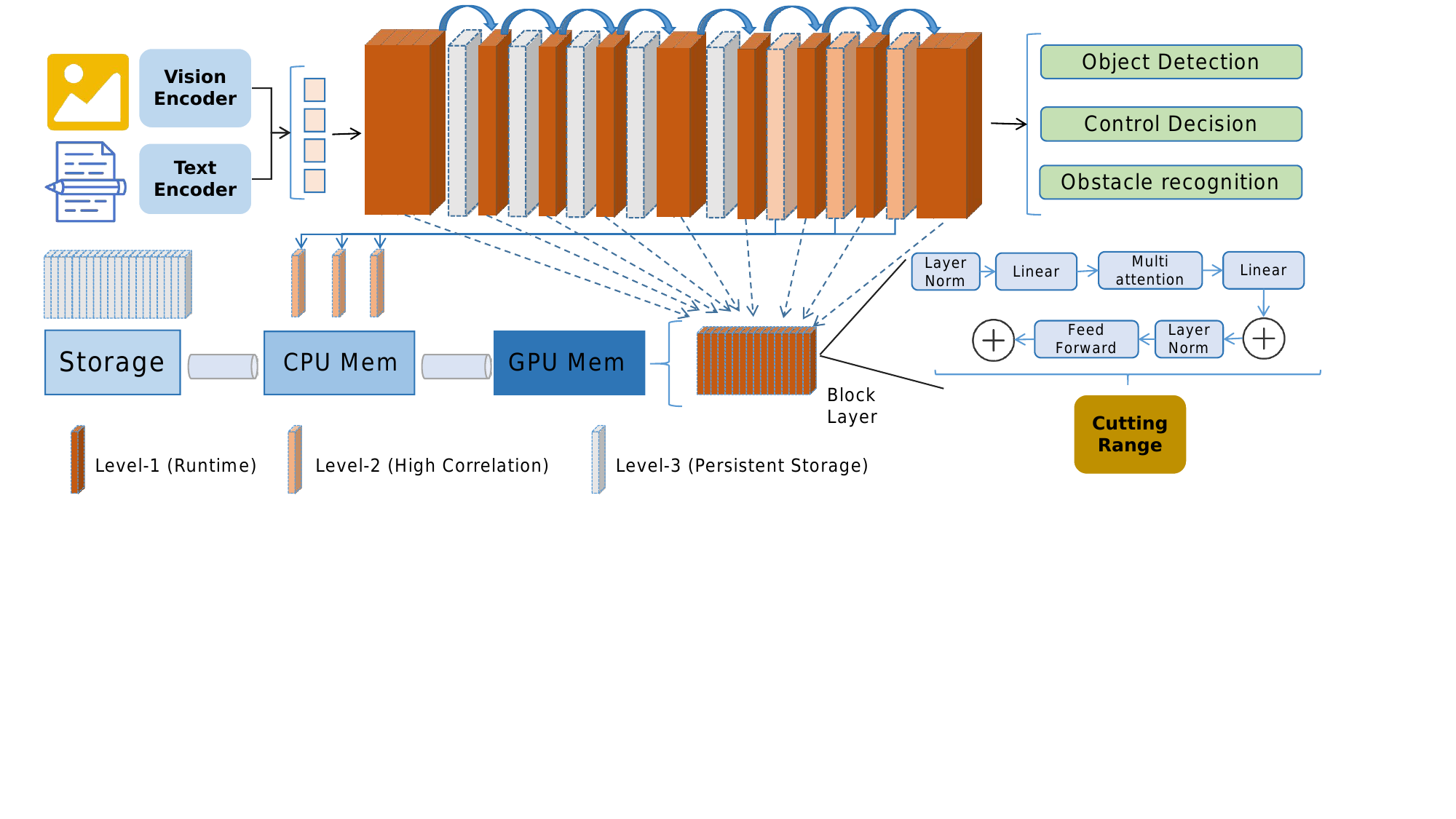}
\vspace{-120pt}
\caption{Overview of the proposed block-layer selective loading framework for multi-modal tasks.
Inputs from the vision and text encoders are processed through a shared Transformer backbone, where task-specific layers are dynamically loaded from storage to GPU memory based on the selected cutting range.}
\vspace{-10pt}
\label{fig:overview}
\end{figure*}

\section{Methodology}
\label{sec:method}

\subsection{System Overview}
\label{sec:overview}
Our goal is to deploy a single large vision-language model to support multiple perception and reasoning tasks on an edge device, while minimizing memory footprint and task-switching latency.
As illustrated in Fig.~\ref{fig:overview}, we adopt a standard multi-model architecture: images and text instructions are encoded by a vision encoder and a text encoder, then fused by a shared Transformer backbone with $N$ blocks, on top of which lightweight task heads are attached for driving-related tasks.
We propose a unified on-demand multi-task sparsity framework with three tightly coupled components:

\textbf{On-demand, fine-grained weight loading for large models.}
We split the model into fine-grained, independently addressable weight units (e.g., block- or layer-level shards) and adopt an on-demand storage mechanism that loads only the required units into GPU memory at inference time, instead of monolithic checkpoints, thereby reducing active memory footprint and enabling fine-grained incremental updates when tasks change.

\textbf{Overlap-aware multi-task sparsity alignment.}
On top of split storage, we introduce a sparsity alignment strategy that encourages different tasks to share as many active units as possible.
By explicitly maximizing structural overlap in the chosen blocks across tasks, we minimize the fraction of parameters that must be swapped during a task switch, making multi-task sparsity compatible with tight memory budgets.

\textbf{Correlation-aware priority pre-loading.}
We further leverage the task-switch patterns illustrated in Fig.~\ref{fig:switching} to rank block layers by transition likelihood. Differential blocks that are associated with frequent task pairs (e.g., Vehicle $\rightarrow$ Traffic Light) are pre-loaded into the CPU cache from disk during ongoing inference. This priority system ensures high-probability blocks are memory-resident, while rarely used ones remain on disk. As a result, storage latency is mostly hidden behind compute time, substantially reducing cold-start delay in dynamic multi-task settings.

\subsection{On-Demand Split Weight for Large Models}
\label{sec:split-storage}
Instead of loading the entire backbone into GPU memory, we partition it into block-level weight shards and treat each shard as an independently addressable unit.
For a given task $t$, only a subset of blocks is required to achieve satisfactory performance.
With one model with totally $N$ layers, we denote by $S_t \subseteq \{0,\dots,N-1\}$ the \emph{skip-layer set} for task~$t$, and by
\[
\mathcal{A}_t = \{0,\dots,N-1\}\setminus S_t
\]
the corresponding set of active blocks.
At runtime, only parameter blocks in $\mathcal{A}_t$ are loaded into GPU memory, while other blocks remain in storage or in CPU memory.
A \emph{cutting range} (Fig.~\ref{fig:overview}) specifies which contiguous segment of the backbone is materialized for a given input, enabling us to dynamically adjust the effective depth of the model.

Let $\Theta = \{\theta_0,\theta_1,\dots,\theta_{N-1}\}$ denote the parameters of the $N$ Transformer blocks in the shared backbone.
Conventional deployment stores and loads $\Theta$ as a monolithic checkpoint.
In contrast, we implement a \emph{split-storage} scheme that decomposes $\Theta$ into block-level shards on secondary storage:
\[
\Theta_{\text{disk}} = \{\theta_k \mid k=0,\dots,N-1\},
\]
where each $\theta_k$ is serialized into an individual file or key-value entry.
On the device, we maintain two caches: a CPU cache $\Theta_{\text{cpu}}$ and a GPU resident set $\Theta_{\text{gpu}}$.
For a given task $t$, the runtime must ensure
\[
\Theta_{\text{gpu}} \supseteq \{\theta_k \mid k \in \mathcal{A}_t\}.
\]

Given a memory budget $B_{\text{gpu}}$, the number of simultaneously active blocks is bounded by
\[
\sum_{k \in \Theta_{\text{gpu}}} \|\theta_k\| \le B_{\text{gpu}},
\]
where $\|\theta_k\|$ denotes the storage size of block $k$ (in parameters or bytes).
When a task switch $t\to t'$ occurs, only the \emph{differential set}
\[
\Delta_{t\to t'} = \mathcal{A}_{t'} \setminus \mathcal{A}_t
\]
must be brought into GPU memory, while blocks in $\mathcal{A}_t \cap \mathcal{A}_{t'}$ are reused.
We first prefetch $\{\theta_k \mid k\in\Delta_{t\to t'}\}$ from disk into $\Theta_{\text{cpu}}$, and then, when GPU memory is available, insert them into $\Theta_{\text{gpu}}$ using a lightweight \texttt{insert} operation that updates only the affected block modules, without reinitializing the entire model.
Concretely, our runtime leverages a CUDA \texttt{insert} interface to directly patch the corresponding block layers into the existing GPU module graph, avoiding teardown and re-creation of the full backbone and further reducing task-switching overhead.

This split-storage design turns sparsity into a first-class system primitive:
block-level activation patterns directly control which shards are loaded, and hence how much I/O is incurred by task switching.
The effectiveness of this mechanism therefore depends critically on how we choose the skip-layer sets $\{S_t\}$.
\subsection{Overlap-Aware Multi-Task Sparsity Alignment}
\label{sec:alignment}

\paragraph{Per-task metric-constrained layer selection.}
For each task $t$, we are given a small calibration set $C_t$ and a task-specific evaluation metric $\mathrm{Metric}_t(\cdot, C_t)$.
Let $s_{\text{full}}^t = \mathrm{Metric}_t(M_{\text{full}}, C_t)$ denote the performance of the full-layer model $M_{\text{full}}$.
We seek a skip-layer set $S_t$ such that the sparse model $M_t$ (obtained by skipping blocks in $S_t$) satisfies a \emph{metric-retention constraint}
\begin{equation}
  \mathrm{Metric}_t(M_t, C_t) \;\ge\; \lambda_t \cdot s_{\text{full}}^t,
  \label{eq:metric-constraint}
\end{equation}
where $\lambda_t\in(0,1]$ is a user-specified retention ratio, which we fix to $0.9$ in all reported results.
We adopt a greedy removal strategy:
starting from the full model, at each iteration we tentatively remove one block $j$ and evaluate the resulting score $s_j$.
Among all candidates that satisfy \eqref{eq:metric-constraint}, we choose the one with the highest score and permanently mark block $j$ as skipped.
This procedure is repeated until either no feasible removable block remains or a desired number of blocks has been removed.
For the first task (Task-1), this greedy selection yields a skip set $S_1$ and an active set $\mathcal{A}_1$ that serve as the initial sparsity pattern.

\paragraph{Shared-pool preference for subsequent tasks.}
For subsequent tasks, we would like their sparse patterns to overlap as much as possible with $\mathcal{A}_1$ (and with each other) so as to reduce $\Delta_{t\to t'}$ during switching.
To this end, we maintain a \emph{shared skip-layer pool} $S_{\text{shared}}$ that collects blocks already skipped by previous tasks.
When constructing the skip set $S_2$ for Task-2, Algorithm~\ref{alg:task2-pref} modifies the greedy removal strategy as follows.

At each removal step, we enumerate all candidate blocks $j$ that are not yet in $S_2$, tentatively remove block $j$, and measure the resulting score $s_j = \mathrm{Metric}_2(M\setminus\{j\}, C_2)$.
We collect feasible candidates whose scores satisfy the metric-retention constraint
\[
s_j \;\ge\; \lambda_2 \cdot s_{\text{full}}^2.
\]
If no candidate is feasible, the procedure stops.
Otherwise, we partition feasible candidates into a shared subset
\[
\mathcal{F}_{\text{shared}} = \{ (j,s_j) \in \mathcal{F} \mid j \in S_{\text{shared}}\}
\]
and a non-shared subset.
If $\mathcal{F}_{\text{shared}}$ is non-empty, we select $j^*$ from $\mathcal{F}_{\text{shared}}$ with the largest score $s_j$; otherwise we fall back to the best candidate from the full set $\mathcal{F}$.
The selected block $j^*$ is added to $S_2$, removed from the model, and the process repeats.
By prioritizing blocks already in the shared pool whenever possible, Algorithm~\ref{alg:task2-pref} encourages Task-2 to skip the same layers as Task-1, thus maximizing overlap in the active sets $\mathcal{A}_1$ and $\mathcal{A}_2$ under the same accuracy constraint.

This idea naturally generalizes to more than two tasks:
whenever a new task $t$ is processed, we build its skip set $S_t$ using the same shared-pool preference, where $S_{\text{shared}}$ aggregates the union of skip sets of previously processed tasks (or a weighted variant if some tasks are considered more important).
In practice, we process tasks in descending order of their priority or traffic volume in the target deployment scenario, so that later, less frequent tasks adapt their sparsity patterns to the dominant ones.
\begin{algorithm}[t]
\caption{Task-2 Layer Selection with Shared-Pool Preference and Metric-Ratio Constraint}
\label{alg:task2-pref}
\DontPrintSemicolon
\KwIn{$M$ (current model), $M_{\text{full}}$ (full-layer reference), $C_2$ (Task-2 calib. set),
$N$ (\# blocks), $n_2$ (\# blocks to remove), $\lambda\in(0,1]$ (retention ratio),
$S_{\text{shared}}$ (shared skip-layer pool)}
\KwOut{$S_2$ (skip-layer set for Task-2)}

$s_{\text{full}} \leftarrow \text{Metric}_2(M_{\text{full}}, C_2)$ \tcp{full-layer baseline}
$S_2 \leftarrow \emptyset$\;

\For{$i \leftarrow 1$ \KwTo $n_2$}{
  $\mathcal{F} \leftarrow \emptyset$ \tcp{feasible candidates}
  \For{$j \leftarrow 0$ \KwTo $N-1$}{
    \If{$j \notin S_2$}{
      $s_j \leftarrow \text{Metric}_2\!\left(M\setminus\{j\},\, C_2\right)$\;
      \If{$s_j \ge \lambda \cdot s_{\text{full}}$}{ add $(j,s_j)$ to $\mathcal{F}$ }
    }
  }
  \If{$\mathcal{F}=\emptyset$}{\textbf{break}}  \tcp{no safe removal}

  \tcp{prefer shared-pool candidates if feasible}
  $\mathcal{F}_{\text{shared}} \leftarrow \{(j,s_j)\in \mathcal{F}\mid j\in S_{\text{shared}}\}$\;
  \eIf{$\mathcal{F}_{\text{shared}}\neq\emptyset$}{
    $(j^*, s^*) \leftarrow \arg\max_{(j,s)\in \mathcal{F}_{\text{shared}}} s$
  }{
    $(j^*, s^*) \leftarrow \arg\max_{(j,s)\in \mathcal{F}} s$
  }

  $S_2 \leftarrow S_2 \cup \{j^*\}$;\quad
  $M \leftarrow \text{remove\_block}(M, j^*)$\;
}
\Return $S_2$\;
\end{algorithm}

\begin{table*}[!htbp]
\centering
\caption{Results on multi-task. 
Acc = accuracy (\%), Sps = sparsity (\% skipped blocks), Lat = latency (s).\vspace{2pt}}
\vspace{-12pt}
\label{tab:main_waymo_multitask}
\renewcommand{\arraystretch}{0.55}
\resizebox{\textwidth}{!}{%
\begin{tabular}{l|ccc|ccc|ccc|ccc|ccc}
\hline
\multirow{2}{*}{Methods} 
& \multicolumn{3}{c|}{Traffic Light} 
& \multicolumn{3}{c|}{Car} 
& \multicolumn{3}{c|}{Obstacle} 
& \multicolumn{3}{c|}{Person} 
& \multicolumn{3}{c}{Bicycle} \\
\cline{2-16}
& Acc & Sps & Lat 
& Acc & Sps & Lat 
& Acc & Sps & Lat 
& Acc & Sps & Lat 
& Acc & Sps & Lat \\
\hline
\multicolumn{16}{c}{\textbf{Qwen2-VL-2B-Instruct}} \\
\hline
Baseline
& 62.3 & 0.0  & 0.062 
& 85.3 & 0.0  & 0.061 
& 64.8 & 0.0  & 0.089 
& 73.9 & 0.0  & 0.061 
& 49.2 & 0.0  & 0.089 \\
SLEB 
& 72.1  & 21.4 & 0.054 
& 100.0 & 25.0 & 0.054 
& 100.0 & 28.6 & 0.073 
& 99.8  & 28.6 & 0.052 
& 100.0 & 28.6 & 0.072 \\
Nav-EE
& 59.3 & 3.6  & 0.059 
& 65.7 & 17.9 & 0.055 
& 59.2 & 17.9 & 0.079 
& 69.5 & 17.9 & 0.056 
& 19.9 & 10.7 & 0.083 \\
\textbf{Ours} 
& \textbf{72.1}  & \textbf{21.4} & \textbf{0.054} 
& \textbf{98.6}  & \textbf{25.0} & \textbf{0.054} 
& \textbf{100.0} & \textbf{28.6} & \textbf{0.073} 
& \textbf{99.5}  & \textbf{25.0} & \textbf{0.054} 
& \textbf{97.6}  & \textbf{21.4} & \textbf{0.076} \\
\hline
\rowcolor{gray!10}
\multicolumn{16}{c}{\textbf{LLaVA-V1.6-Vicuna-7B}} \\
\hline
\rowcolor{gray!10}
Baseline
& 69.7 & 0.0 & 0.086 
& 82.1 & 0.0 & 0.129 
& 80.7 & 0.0 & 0.175 
& 75.3 & 0.0 & 0.089 
& 37.2 & 0.0 & 0.140 \\
\rowcolor{gray!10}
SLEB
& 75.2 & 31.3 & 0.064 
& 100.0 & 43.8 & 0.079 
& 99.7 & 37.5 & 0.119 
& 99.8 & 34.4 & 0.064 
& 88.2 & 46.9 & 0.091 \\
\rowcolor{gray!10}
Nav-EE
& 98.2 & 21.9 & 0.070 
& 67.0 & 31.3 & 0.093 
& 98.0 & 21.9 & 0.145 
& 91.0 & 31.3 & 0.066 
& 10.5 & 9.4  & 0.135 \\
\rowcolor{gray!10}
\textbf{Ours} 
& \textbf{99.2} & \textbf{46.9} & \textbf{0.053} 
& \textbf{100.0} & \textbf{46.9} & \textbf{0.076} 
& \textbf{91.6} & \textbf{50.0} & \textbf{0.101} 
& \textbf{100.0} & \textbf{50.0} & \textbf{0.052} 
& \textbf{99.6} & \textbf{50.0} & \textbf{0.087} \\
\hline
\end{tabular}
}
\vspace{-16pt}
\begin{flushleft}
\end{flushleft}
\end{table*}

\begin{figure*}[t]
    \centering
    \begin{subfigure}[b]{0.24\textwidth}
        \centering
        \includegraphics[width=\linewidth]{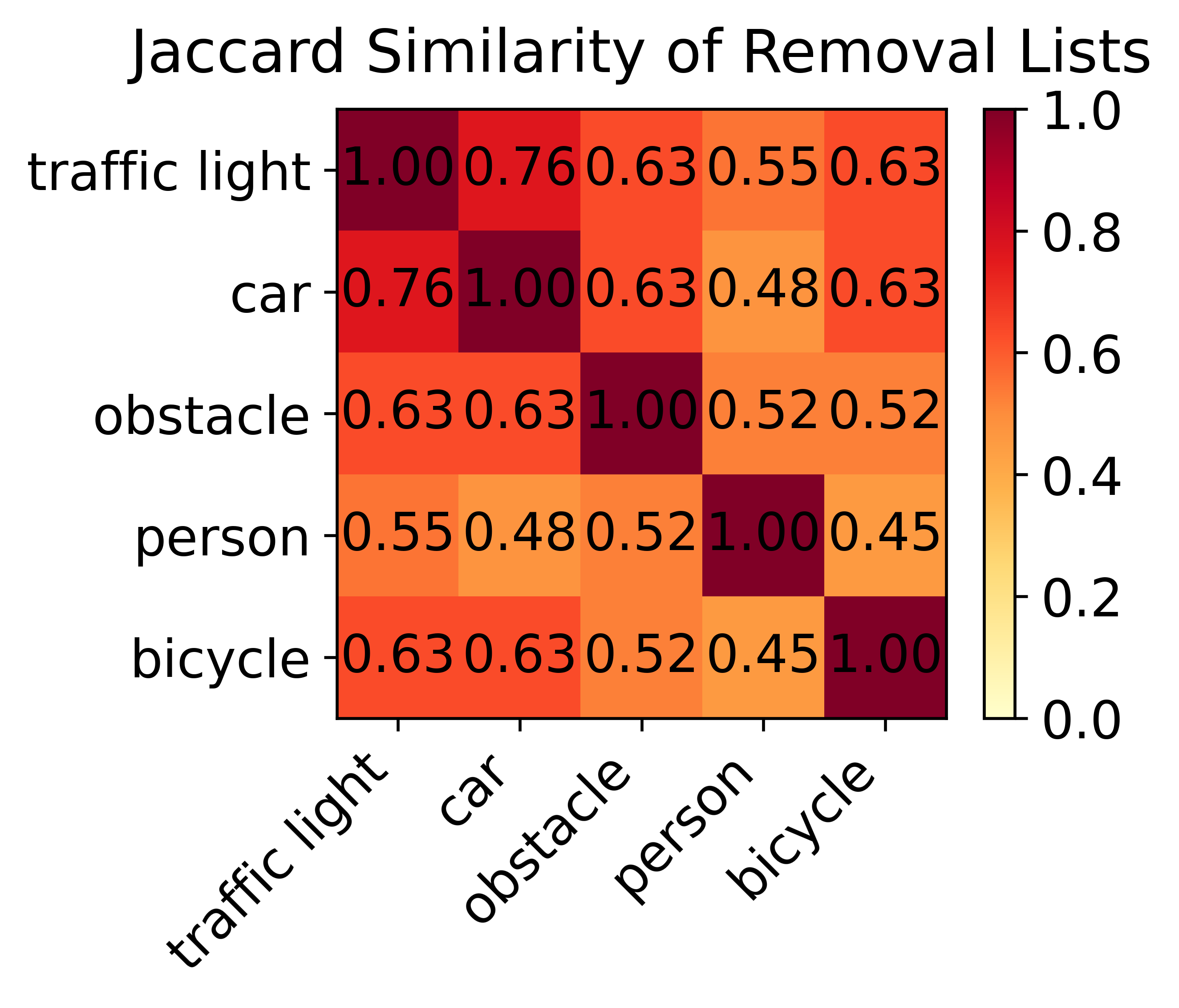}
        \vspace{-20pt}
        \caption{LLaVA-7B / Ours}
        \label{fig:jaccard_a}
    \end{subfigure}
    \begin{subfigure}[b]{0.24\textwidth}
        \centering
        \includegraphics[width=\linewidth]{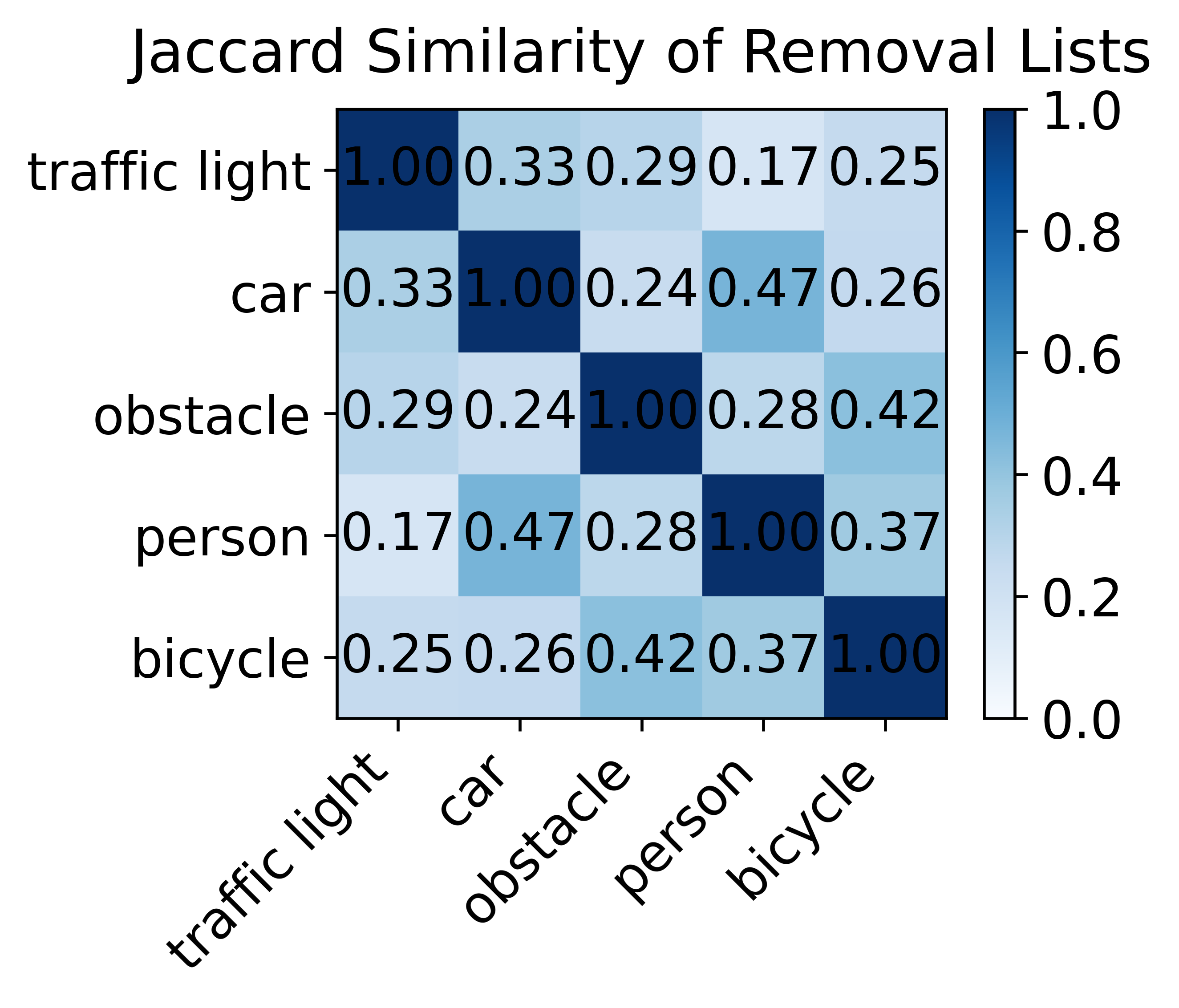}
        \vspace{-20pt}
        \caption{LLaVA-7B / SLEB}
        \label{fig:jaccard_b}
    \end{subfigure}
    \begin{subfigure}[b]{0.24\textwidth}
        \centering
        \includegraphics[width=\linewidth]{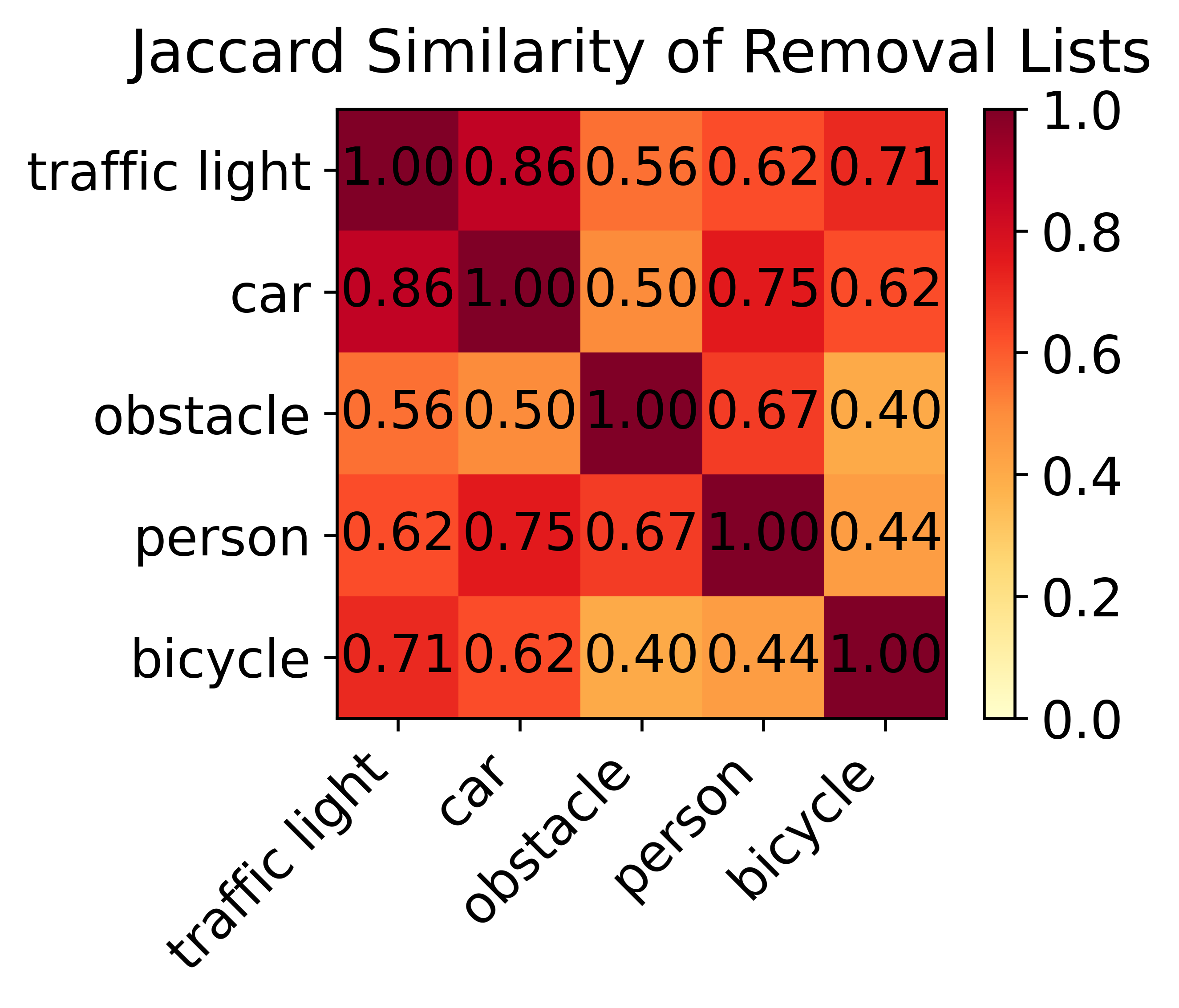}
        \vspace{-20pt}
        \caption{Qwen2-VL-2B / Ours}
        \label{fig:jaccard_c}
    \end{subfigure}
    \begin{subfigure}[b]{0.24\textwidth}
        \centering
        \includegraphics[width=\linewidth]{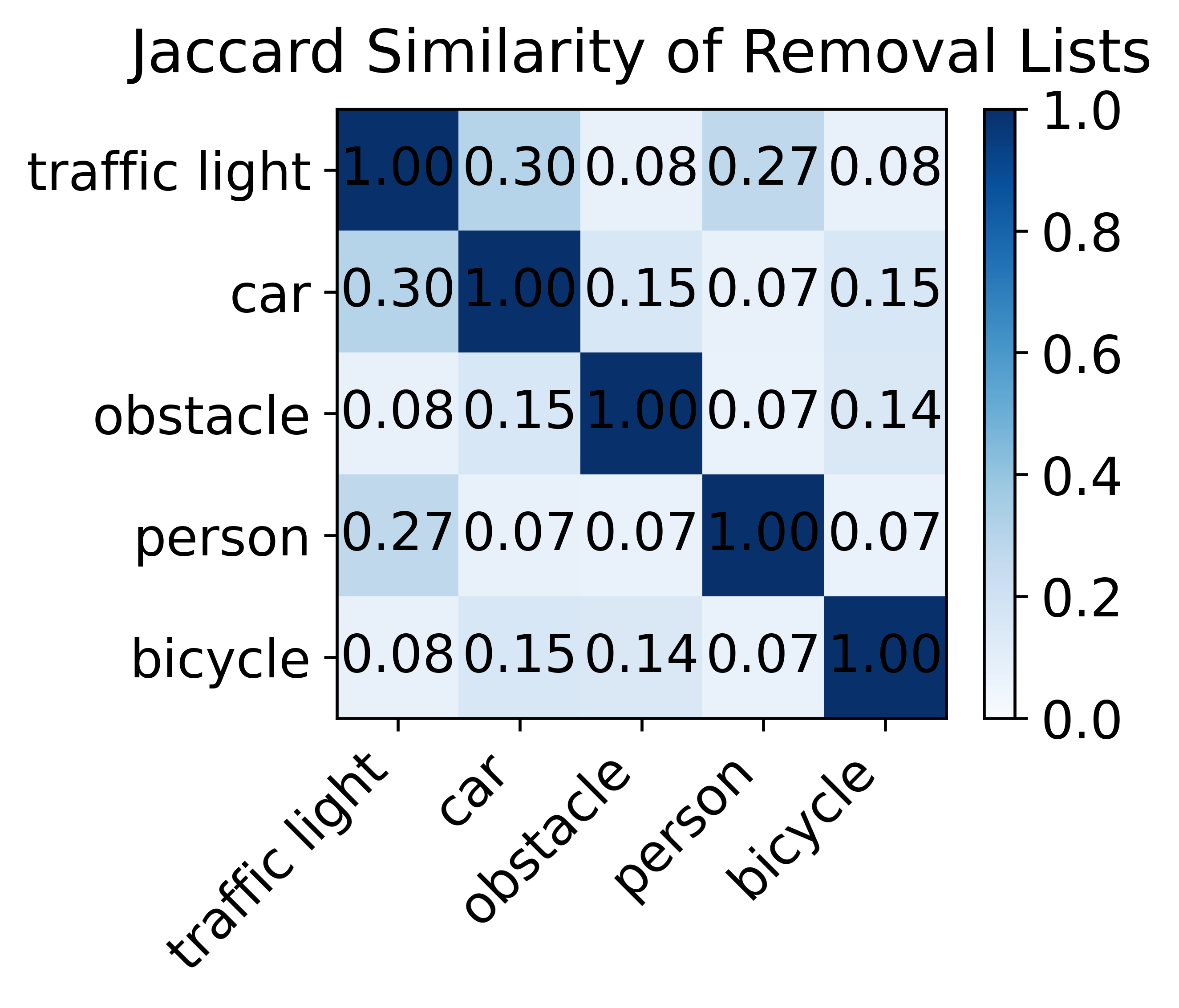}
        \vspace{-20pt}
        \caption{TQwen2-VL-2B / SLEB}
        \label{fig:jaccard_d}
    \end{subfigure}
    \vspace{-10pt}
    \caption{Pairwise Jaccard similarity heatmaps for different task settings.}
    \label{fig:jaccard_all}
\end{figure*}
\subsection{Correlation-Aware Priority Pre-Loading}
\label{sec:scheduler}

Even with sparsity alignment, switching from task \( t \) to \( t' \) still requires loading the differential set \( \Delta_{t \to t'} = \mathcal{A}_{t'} \setminus \mathcal{A}_t \) from storage. To mitigate this latency, we leverage temporal transition patterns (cf. Fig.~\ref{fig:switching}) and propose a correlation-aware \textit{priority} pre-loading strategy guided by task transition statistics.

\noindent\textbf{Task transition--guided priority tiers.}
From logged sequences (e.g., nuScenes trainval), we estimate a first-order transition matrix
\[
P(t' \mid t) = \frac{\#(t \to t')}{\sum_u \#(t \to u)},
\]
and extract the top-$K$ likely successors $\mathcal{N}(t)$.
Based on this, we assign block layers into three hierarchical tiers:
Level-1 contains the runtime blocks used by the current task $t$ and kept in GPU memory;
Level-2 contains blocks belonging to highly correlated next tasks $t' \in \mathcal{N}(t)$, which are prioritized for pre-loading into CPU memory;
Level-3 comprises the remaining blocks that stay on SSD/HDD.
This priority scheme lets likely-needed blocks be staged on CPU in advance, hiding most disk latency when task switching occurs.

\noindent\textbf{Pre-loading policy.}
Given a CPU-side memory budget $B_{\text{cpu}}$, the system greedily fills it with Level-2 blocks while $t$ is running. Upon switching to $t'$, if \( \Delta_{t \to t'} \) is already staged, only fast CPU-to-GPU transfers are needed; otherwise, slower disk loading is triggered.

This correlation-aware mechanism introduces negligible scheduling overhead but significantly reduces real-time switching latency across diverse multi-task streams.

\subsection{Real-World Deployment and Experimental Setup}
\noindent\textbf{Test Platform.}  
We deploy our method on a real autonomous vehicle equipped with an industrial PC running an NVIDIA RTX~3090 GPU. The vehicle uses the open-source \emph{Autoware.Universe} stack, where our VLM module is integrated into the perception pipeline to provide structured semantic cues—including object types, spatial interactions, and traffic signal understanding—based on front-view camera input.
\begin{figure}[H]
  \centering
  \vspace{-12pt}
  \includegraphics[width=0.5\textwidth,trim=1cm 12.5cm 0cm 0cm, clip]{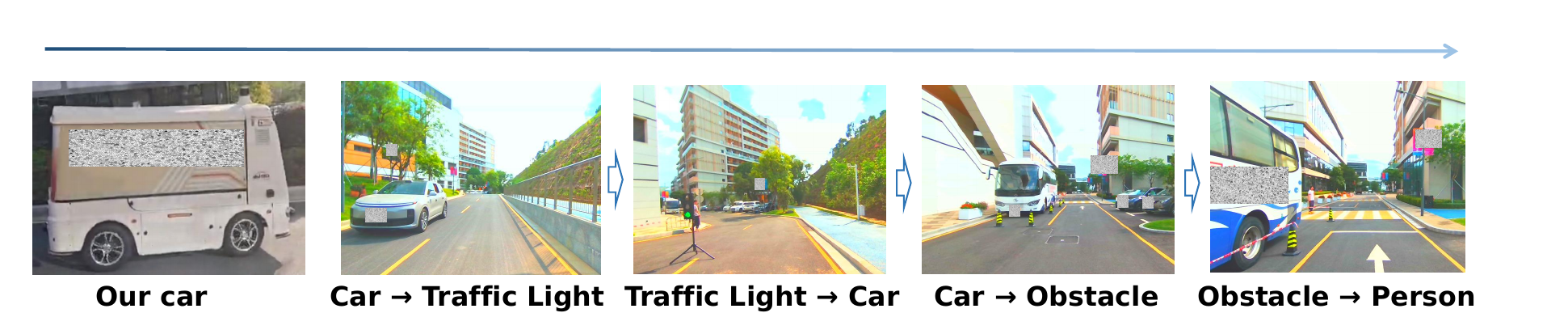}
  \caption{Execution order of perception tasks on our autonomous vehicle, with each image illustrating a key scene in the sequence (Car → Traffic Light → Car → Obstacle → Person).}
  \vspace{-10pt} 
  \label{fig:real}
\end{figure}

\noindent\textbf{Scenario}  
To evaluate the system under practical conditions, we design a driving route that sequentially triggers five major perception tasks: Vehicle Detection, Lighting Decision, Obstacle Detection (e.g., cones), and Person Detection. As illustrated in Fig.~\ref{fig:real}, the sequence includes recurring transitions such as \emph{Car → Traffic Light}, \emph{Traffic Light → Car}, \emph{Car → Obstacle}, and \emph{Obstacle → Person}, with the VLM dynamically adjusting its inference path as task conditions shift.



\section{Experiments}
\subsection{Experimental Setup}
\noindent\textbf{Datasets and tasks.}
We build a multi-dataset driving benchmark using the \emph{nuScenes} trainval split as the main source of real driving scenes and trajectories.
Because nuScenes lacks explicit obstacle and traffic-light labels, we augment it with \emph{CODA} for generic obstacles and \emph{Bosch Small Traffic Lights} for signal states.
These datasets are unified into five perception tasks—\emph{Traffic Light}, \emph{Car}, \emph{Obstacle}, \emph{Person}, and \emph{Bicycle} detection/recognition—used for both per-task evaluation and multi-task switching experiments on edge devices.

\noindent\textbf{Model configuration.}
We use two representative vision–language backbones, \textbf{LLaVA} and \textbf{Qwen2}, each evaluated in a full-layer baseline and our sparse on-demand configuration.

\noindent\textbf{Baselines.}
We compare our framework with two representative sparsity methods: \emph{SLEB}~\cite{24icml-sleb}, which eliminates redundant transformer blocks via task-specific sparsity, and \emph{Nav-EE}~\cite{25arxiv-Nav-EE}, which uses navigation-guided, task-specific early-exit layers for VLMs.

\noindent\textbf{Metrics.}
We report both task-level and system-level performance.
Task-level metrics include accuracy (Acc), end-to-end per-frame inference latency (I/O + compute), and sparsity (fraction of skipped transformer blocks); system-level metrics include task-switch overhead (time during transitions between heterogeneous tasks) and GPU utilization under dynamic task streams.

\subsection{Multi-Task Perception Results}
\label{sec:exp_waymo}
Table~\ref{tab:main_waymo_multitask} reports per-task accuracy (Acc), sparsity (Sps) and inference latency (Lat) for the five tasks.
Across both Qwen2-VL-2B--Instruct and LLaVA-V1.6-Vicuna-7B, our method achieves the best overall trade-off: it preserves or improves accuracy while significantly increasing sparsity and reducing inference latency.

For \textbf{Qwen2-VL-2B}, the full model exhibits modest average accuracy (67.1\%) with no sparsity.
SLEB greatly boosts accuracy but requires strong task-specific sparsity patterns, and Nav-EE suffers from severe accuracy drops on certain tasks (e.g., Bicycle at 19.9\%).
In contrast, \textbf{Ours} reaches 93.56\% average accuracy, comparable to or better than SLEB on all tasks, while maintaining 24.29\% average skipped blocks and keeping latency comparable to SLEB.

For \textbf{LLaVA-7B}, the baseline already attains reasonably high accuracy on most tasks except Bicycle (37.2\%), but with no sparsity and the highest latency.
SLEB further improves accuracy yet only moderately increases sparsity, whereas Nav-sEE yields very uneven behavior (e.g., Bicycle accuracy drops to 10.5\%).
\textbf{LLaVA-7B Ours} delivers the strongest overall performance: it achieves 98.08\% average accuracy, pushes sparsity to 48.75\% on average, and reduces per-task latency to 73.69\,ms, substantially outperforming both SLEB and Nav-EE in the latency–accuracy–sparsity trade-off.

These results confirm that our on-demand split storage and overlap-aware multi-task sparsity alignment can sustain high perception quality across heterogeneous tasks while aggressively skipping blocks, providing a solid foundation for subsequent analysis of task-switching efficiency.
\begin{figure}[t]
  \centering
  \includegraphics[width=0.4\textwidth,trim=0cm 0cm 18cm 0cm, clip]{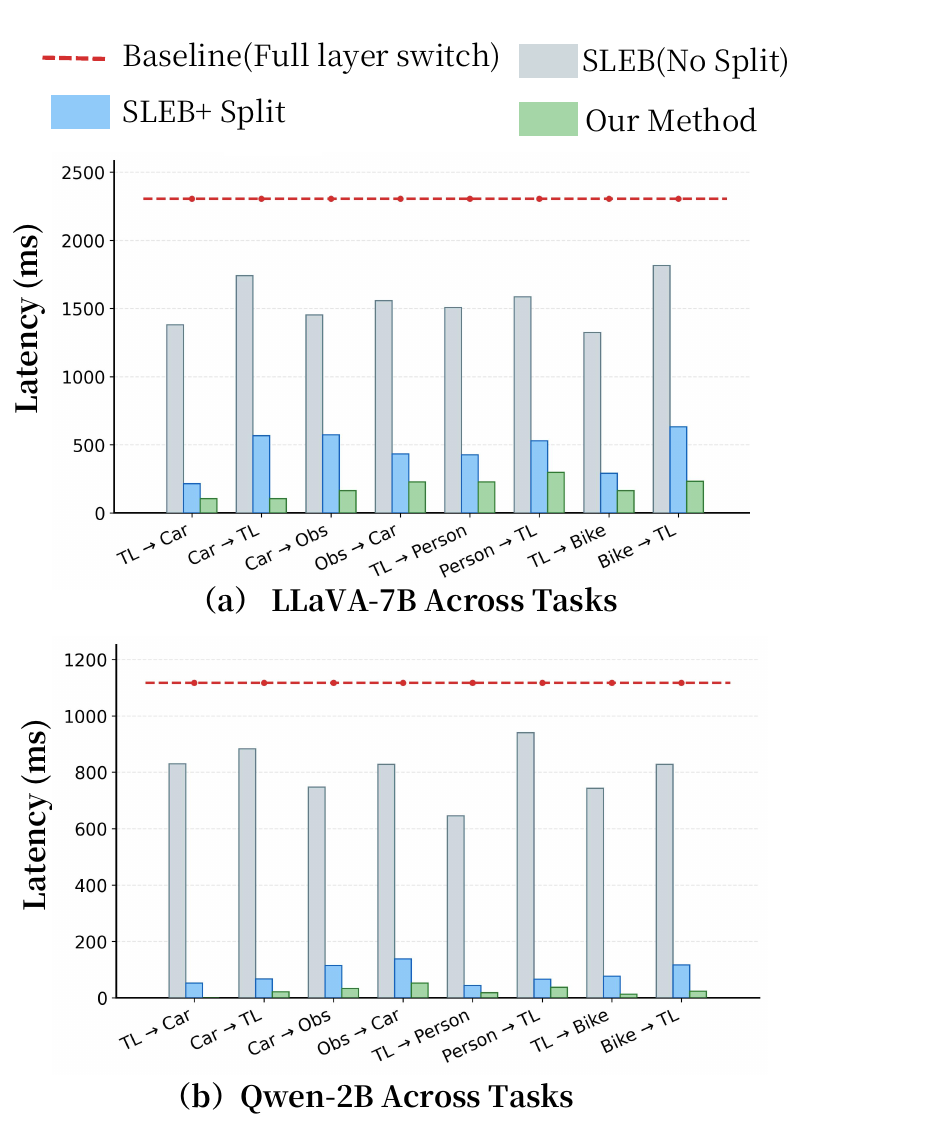}
  \caption{Task-switching latency.}
  \vspace{-20pt} 
  \label{fig:latency}
\end{figure}

\subsection{Analysis of Multi-Task Sparsity Alignment}
\label{sec:exp_alignment}

To better understand how our framework reduces task-switching overhead, we analyze the overlap of skip-layer patterns across the five perception tasks.
Fig.~\ref{fig:jaccard_all} visualizes the pairwise Jaccard similarity of removal lists (skip-layer sets) for LLaVA-7B and Qwen2-VL-2B under our method and SLEB.
Each heatmap entry measures the fraction of Transformer blocks that are skipped by both tasks, i.e., the structural overlap of their sparse configurations.

For \textbf{LLaVA-7B}, our method (Fig.~\ref{fig:jaccard_all}a) yields consistently high off-diagonal similarities (mostly $>0.6$) among tasks, especially for high-frequency pairs such as Traffic Light–Car and Car–Person.
In contrast, SLEB (Fig.~\ref{fig:jaccard_all}b) produces much lower overlaps (often $<0.3$), because each task independently learns its own sparsity pattern without considering cross-task alignment.
A similar trend appears on \textbf{Qwen2-VL-2B}: our method (Fig.~\ref{fig:jaccard_all}c) again exhibits strong sharing of removal lists across tasks, whereas SLEB (Fig.~\ref{fig:jaccard_all}d) shows almost orthogonal patterns.

Fig.~\ref{fig:latency} reports end-to-end task-switch latency for representative task pairs. Baseline reloads full layers, SLEB applies sparsity without split, SLEB+Split uses our split storage, and Our Method combines split storage with multi-task alignment to achieve the lowest latency.
Each group corresponds to a transition between two of the five perception tasks
(\emph{Traffic Light} (TL), \emph{Car}, \emph{Obstacle}(Obs), \emph{Person}, \emph{Bicycle}(Bike)),
and we compare four configurations:
the \emph{Baseline} full-layer model (red dashed line),
\emph{SLEB (No Split)} with task-wise sparsity but monolithic checkpoints,
\emph{SLEB+Split} which uses only our split-storage mechanism,
and \emph{Our Method} which activates all three components.

For \textbf{LLaVA-7B} (Fig.~\ref{fig:latency}a), the Baseline and SLEB (No Split) both require roughly $2.4$\,s per switch, as every transition reloads a full (sparse) checkpoint.
Adding split storage (\emph{SLEB+Split}) already cuts latency to about $400$–$600$\,ms by loading only the used blocks.
Our full framework further reduces switch time to roughly $100$–$300$\,ms, thanks to the high overlap of active blocks in Fig.~\ref{fig:jaccard_all} and the small differential sets $\Delta_{t\to t'}$.

A similar pattern appears for \textbf{Qwen2-VL-2B} (Fig.~\ref{fig:latency}b): the Baseline remains around $1.1$\,s per switch, SLEB (No Split) offers little improvement, while \emph{SLEB+Split} brings latency into the $100$–$200$\,ms range.
\emph{Our Method} further pushes many switches down to tens of milliseconds, and in one task pair the measured latency is effectively $0$\,ms because the next task only needs to \emph{drop} a single block with no CPU$\to$GPU transfer.

Overall, sparsity alone is insufficient for efficient multi-task deployment: without storage-aware design and cross-task alignment, SLEB still pays the cost of full-checkpoint swapping.
By combining split storage, overlap-aware multi-task sparsity, and correlation-aware scheduling, our method turns task switches into small incremental updates and achieves an order-of-magnitude reduction in switching latency.

\begin{table}[t]
\centering
\caption{Performance comparison of data transport and task switching on Real Car platform(LaVA-V1.6-Vicuna-7B).}
\vspace{-10pt}
\label{tab:transport_switch}
\renewcommand{\arraystretch}{0.6} 
\resizebox{\linewidth}{!}{
\begin{tabular}{l|l|c|c}
\hline
Method & Scenario & Latency (ms) & GPU Util (MiB) \\
\hline
Baseline & Task Switch (full) & 1566.5 & 13472.43 \\
\hline
\multirow{4}{*}{SLEB} 
& Car → Traffic Light & 1036.3 & 8068.21 \\
& Traffic Light → Car & 940.4 & 9612.27 \\
& Car → Obstacle & 1118.5 & 8068.21 \\
& Obstacle → Person & 1013.3 & 8840.24 \\
\hline
\multirow{4}{*}{SLEB+Spilt} 
& Car → Traffic Light & 430.2 & 8068.21 \\
& Traffic Light → Car & 190.9 & 9612.27 \\
& Car → Obstacle & 466.2 & 8068.21 \\
& Obstacle → Person & 361.6 & 8840.24 \\
\hline
\multirow{4}{*}{Ours} 
& Car → Traffic Light & \textbf{117.1} & \textbf{7682.20} \\
& Traffic Light → Car & \textbf{96.1} & \textbf{7682.20} \\
& Car → Obstacle & \textbf{164.4} & \textbf{7682.20} \\
& Obstacle → Person & \textbf{240.6} & \textbf{7296.18} \\
\hline
\end{tabular}
} 
\vspace{-20px}
\end{table}
\subsection{Real-Car Multi-Task Switching Results}
\label{sec:exp_real_car}

Table~\ref{tab:transport_switch} reports task-switching latency and GPU memory utilization (GPU Util) of different methods on the real-car platform, with LLaVA-V1.6-Vicuna-7B as the base model. The experiment targets practical autonomous driving switching scenarios (e.g., Car→Traffic Light, Obstacle→Person) to verify on-board efficiency.

For latency, our method achieves the lowest performance (96.1 – 240.6 ms), outperforming the full-model Baseline (1566.5 ms), SLEB (940.4 – 1118.5 ms), and SLEB+Spilt (190.9 – 466.2 ms). This gain comes from our correlation-aware priority pre-loading that minimizes data transfer overhead.

The GPU Util optimization is a key highlight for on-board systems. The Baseline consumes 13472.43 MiB (limiting other critical tasks like path planning), while SLEB and SLEB+Spilt still use 8068.21–9612.27 MiB. Our method reduces utilization to 7296.18–7682.20 MiB (up to 46\% lower than Baseline) via hierarchical block management: only high-priority (runtime, high-correlation, shared) blocks are retained in GPU memory, avoiding redundant occupation.

This low GPU utilization avoids memory bottlenecks, reduces resource competition, and mitigates thermal throttling risks in harsh on-vehicle environments. Real-car results confirm our method balances latency and GPU efficiency, providing a reliable solution for autonomous driving perception systems.
\section{Conclusion}
We presented a switching-aware sparsity framework for multi-task deployment of large vision–language models on edge devices.
By combining on-demand fetching, overlap-aware multi-task sparsity alignment, and correlation-aware priority pre-loading, our approach greatly reduces task-switch latency while preserving or even improving accuracy.
Experiments on Nuscenes-style multi-task perception and real-vehicle deployment show consistent gains over SLEB, and early-exit baselines, highlighting the potential of system-level sparsity design for practical large-model deployment.

\bibliographystyle{ACM-Reference-Format}
\bibliography{sample-base}
\end{document}